\documentclass[10pt, a4paper]{article}

\usepackage{lrec-coling2024} 
\usepackage{booktabs}
\usepackage{graphicx}
\usepackage{comment}
\usepackage{multirow}
\usepackage[inline]{enumitem}

\title{Extracting Biomedical Entities from Noisy Audio Transcripts}

\name{Nima Ebadi$^1$, Kellen Morgan$^2$, Adrian Tan$^3$, Billy Linares$^3$, Sheri Osborn$^4$,\\ {\bf \large{Emma Majors$^5$, Jeremy Davis$^5$, and Anthony Rios$^4$}}} 

\address{$^1$Department of Electrical and Computer Engineering, The University of Texas at San Antonio\\$^2$Department of Management Science and Statistics, The University of Texas at San Antonio\\$^3$Data Analytics, The University of Texas at San Antonio\\$^4$Department of Information Systems and Cyber Security, The University of Texas at San Antonio\\ $^5$Department of Neurology, Division of Neuropsychology, UT Health San Antonio \\
         kellen.morgan@my.utsa.edu, emma.majors06@gmail.com, davisj20@uthscsa.edu\\
         \{nima.ebadi, billy.linares, sheri.orborn, anthony.rios\}@utsa.edu\\}

\abstract{
Automatic Speech Recognition (ASR) technology is fundamental in transcribing spoken language into text, with considerable applications in the clinical realm, including streamlining medical transcription and integrating with Electronic Health Record (EHR) systems. Nevertheless, challenges persist, especially when transcriptions contain noise, leading to significant drops in performance when Natural Language Processing (NLP) models are applied. Named Entity Recognition (NER), an essential clinical task, is particularly affected by such noise, often termed the ASR-NLP gap. Prior works have primarily studied ASR's efficiency in clean recordings, leaving a research gap concerning the performance in noisy environments. This paper introduces a novel dataset, BioASR-NER, designed to bridge the ASR-NLP gap in the biomedical domain, focusing on extracting adverse drug reactions and mentions of entities from the Brief Test of Adult Cognition by Telephone (BTACT) exam. Our dataset offers a comprehensive collection of almost 2,000 clean and noisy recordings. In addressing the noise challenge, we present an innovative transcript-cleaning method using GPT4, investigating both zero-shot and few-shot methodologies. Our study further delves into an error analysis, shedding light on the types of errors in transcription software, corrections by GPT4, and the challenges GPT4 faces. This paper aims to foster improved understanding and potential solutions for the ASR-NLP gap, ultimately supporting enhanced healthcare documentation practices.
 \\ \newline \Keywords{Named Entity Recognition, Biomedical Informatics, Audio Speech Recognition} }

\begin{document}

\maketitleabstract

\section{Introduction}

Automatic Speech Recognition (ASR) technology is pivotal in converting spoken language into written text and finds critical applications within clinical contexts. One important use is expediting medical transcription processes and efficiently documenting doctor-patient interactions. This seamless conversion reduces the time and resources traditionally spent on manual transcription, affording healthcare professionals more time for focused patient care. Specifically, ASR can seamlessly integrate into Electronic Health Record (EHR) systems, enabling real-time dictation of diagnoses, treatment plans, and patient notes, thereby augmenting the accuracy and immediacy of clinical documentation. Hence, this technology holds substantial promise in revolutionizing healthcare documentation practices. After successful conversion from audio to text, natural language processing (NLP) tools can be applied to the transcriptions for various tasks~\cite{szymanski2023aren}. Unfortunately, transcription is not accurate, particularly in noisy environments. Moreover, when NLP models are applied to noisy data that does not match the training data distribution, large drops in performance may be observed.

This paper focuses on the biomedical NLP entity recognition (NER) task applied to noisy audio transcripts. Named entity recognition is vital for many important clinical tasks, from extracting social determinants of health mentions from clinical notes to extracting mentions of adverse drug reactions. Clinicians may not be able to capture everything stated to them by a patient (e.g., specific adverse reactions to a drug), particularly if they need to transcribe information after an interaction via rote memory. Hence, if ASR can be used to record patient-clinician interactions, then NER systems can be applied to extract clinically relevant information for later use. We explore the viability of NER systems applied to noisy transcripts to better understand their performance in real-world settings, where records may have multiple speakers and background sounds.

\citet{szymanski2023aren} has recently called this difference in performance the ASR-NLP gap. At a high level, there are two primary causes for the ASR-NLP gap. First, transcription errors can completely change the words mentioned. For instance, if someone mentions the word ``headache'' (which could be a mention of a drug side effect), but if it is recognized as ``headway,'' then a traditional NER system would be unable to identify it. Second, the data distribution changes. Models trained on clean, non-transcribed data may capture different patterns in the text that are not available in the transcribed text. The patterns may be as simple as differences in punctuation and capitalization, but such patterns are essential for accurate NER. 

Much of the prior work on studying ASR systems, particularly in biomedical domains, has focused on either developing or evaluating ASR systems for novel patient populations~\cite{tran2023mm} or training and evaluating NLP systems on carefully corrected and relatively clean transcripts. For work evaluating ASR systems in the clinical domain, there have been low word error rates (WER) reported (e.g., 11\%~\cite{tran2023mm} 24.3\%~\cite{hacking2023development}, and 10\%~\cite{king2023voice}). However, the studies often report results on relatively clean recordings (e.g., without multiple background speakers or substantial background noise). Sometimes transcripts that are very noisy are completely removed from the evaluation data~\cite{king2023voice},  potentially resulting in overly optimistic performance.  Prior works have reported WERs much worse than the reported numbers in the clinical setting~\cite{kodish2018systematic}, with WERs in the range of 30\% to 60\%. Moreover, \citet{kodish2018systematic} also evaluated concept extraction software on transcriptions. However, they did not compare the performance difference between clean data and noisy transcripts. The numbers are still generally reported on ``clean'' transcripts with minimal background noise and background speakers. Finally, they do not provide any natural next steps for improving performance. Hence, the results may be much worse when evaluating substantially noisy environments.

In this paper, we develop a new dataset so biomedical NLP researchers can directly improve and explore the biomedical ASR-NLP gap. Specifically, we introduce a dataset that extracts adverse drug reaction mentions and a dataset that extracts fruits and animals that would be mentioned as part of the Brief Test of Adult Cognition by Telephone (BTACT) exam. To the best of our knowledge, this will be the first publicly available dataset to allow for careful evaluation of the ASR-NLP gap in the biomedical domain. 




In summary, based on current research gaps in the ASR-NLP gap for biomedical applications, this paper makes the following contributions:
\begin{enumerate}[label=\bfseries(\roman*)]
    \item We introduce a novel dataset of nearly 2000 clean and noisy recordings for biomedical-related ASR-NER called BioASR-NER.\footnote{The dataset is available at \url{https://zenodo.org/records/10864063}.}
    \item We introduce a simple approach to improving model performance via a transcript-cleaning procedure using GPT4. We explore both zero-shot and few-shot methodologies for when ground-truth noisy and cleaned transcription pairs are limited.
    \item Finally, we perform an informative error analysis showcasing the types of errors made by the transcription software, the type of errors GPT4 corrects, and the types of errors GPT4 cannot handle accurately.
\end{enumerate}

\section{Related Work}

In this work, we describe two major research lines relevant to this paper: Biomedical ASR-NLP, which includes work on Biomedical ASR technologies and NLP applied to transcriptions (clean and noisy if available), and Biomedical NER, which discusses some recent work on developing methods to extract biomedical entities from text.

\subsection{Biomedical NER}
There have been many datasets and methods developed for the detection of biomedical entities~\cite{leaman2008banner,song2021deep,rocktaschel2012chemspot,chiu2021recognizing,lee2020biobert,sun2021deep,lopez2021combining,weber2021hunflair}. Specifically, there are biomedical NER tasks including, but not limited to, extracting mentions of social determinants of health from electronic medical records, detecting adverse drug interactions in patient self-reports~\cite{karimi2015cadec}, extracting chemical and drugs mentions~\cite{rocktaschel2012chemspot}, and extracting gene mentions in biomedical research articles~\cite{pyysalo2007bioinfer}.

Many novel methodological approaches have been developed for each of the tasks. For example, \citet{lee2020biobert} developed a specialized BERT model tailored for biomedical applications, demonstrating improvements over previous state-of-the-art results. Additionally, HunFlair~\cite{weber2021hunflair} introduced a methodology that combines word, contextual, and character embeddings within a unified framework, achieving state-of-the-art performance.  \citet{tong2021multi} introduce a multi-task learning framework for biomedical NER that integrates multiple related training objectives to improve entity extraction. Similarly, \citet{watanabe2022auxiliary} improve biomedical NER by incorporating auxiliary learning with multiple datasets. \citet{guan2023prefix} incorporated information between word pairs to improve biomedical NER performance. And more recently, \citet{ghosh2023bioaug} explored synthetic data augmentation to improve low-resource biomedical NER. Similarly, \citet{chen2023few} improved few-shot NER via contrastive prompt tuning.

Overall, our work is most related to research on out-of-domain performance of information extraction systems~\cite{rios2018generalizing,jia2019cross,poerner2020inexpensive,vu2020effective,nguyen2022hardness}. \citet{poerner2020inexpensive} train word embeddings on the target domain and the align them to the general domain to improve generalization. \citet{nguyen2022hardness} introduce ``hardness'' related information to better generalize biomedical NER models across domains. However, contrary to prior research, our work differs in one major way. Specifically, we are focused on a particular kind of domain shift. Prior work has explored two disparate domains such as social media and electronic health records. In our paper, the underlying data does not change. Instead, the style of the content changes because of the noisy channel caused by the transcription process. 

\begin{figure*}
    \centering
    \includegraphics[width=\linewidth]{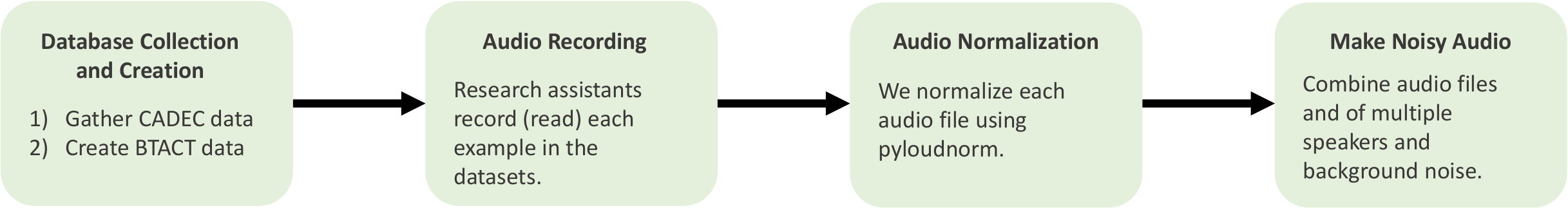}
    \caption{Overview of our data collection process. The process has four main steps: 1) We collect the initial datasets (CADEC and BTACT); 2) Graduate assistants read and record the text in the datasets; 3) We normalize each audio file to the same loudness; and 4) we generate noisy audio files by merging multiple speakers and adding background noise.}
    \label{fig:dproc}

\end{figure*}

\subsection{Biomedical ASR-NLP}
As discussed in the Introduction, much of the work on automatic speech recognition (ASR) systems provides overly optimistic word error rates (WER). Many datasets lack background noise and only have a single speaker. Yet, real-world datasets may have background noise, multiple background speakers of various volumes, and even dropped connections. Recent studies and reviews have discussed how digital scribes (ASR systems) are necessary to reduce physician burden to provide more reliable care~\cite{quiroz2019challenges,van2021digital}.

Recently, there have been two major research directions for biomedical applications related to ASR. First, new ASR systems have been proposed directly for particular patient populations~\cite{kodish2018systematic}. For example, \citet{hacking2023development} introduces a novel ASR system for older adults in an interview setting. Likewise, there has been substantial work on developing and improving biomedical ASR systems for languages besides English~\cite{dhuriya2022improving}. Second, there has been research that has evaluated commercial ASR systems in the biomedical domain. For example, \cite{tran2023mm} evaluated proprietary ASR systems for their ability to detect non-lexical conversational sounds such as ``Mm-hm'' and ``Uh-uh'', which can be clinically relevant in many scenarios. The authors found that current systems are unable to detect them regularly. Likewise, \citet{paats2015evaluation} evaluated ASR systems in Estonian languages. Finally, there has been research that has developed and evaluated NLP systems on ASR transcripts. For example, \cite{ganoe2021natural} develop NER tools to extract medication mentions in transcripts of primary care conversations. Yet, much of the prior work applying NLP tools to transcripts has used ``cleaned'' transcripts with limited transcription errors where a human has ensured the transcript is accurate. In this work, we focus on noisy transcripts in the presence of background noise and multiple speakers. Moreover, for work that evaluates ASR systems using WER, that performance does not correlate with the quality of the transcription by a human evaluator and does not correlate with downstream performance on NLP tasks~\cite{whetten2023evaluating,szymanski2023aren}.

From a methodological standpoint, some recent work has explored reducing transcription errors. To this end, our work is similar to \citet{mani2020towards} that developed a seq2seq method to reduce transcription errors applied after a mainstream ASR process. Our work expands on this direction in two ways. First, we provide a unique dataset in a domain that lacks publicly available data. Second, our work analyzes the impact on NER directly, not WER, which can negatively correlate with NER performance.

Overall, our work is most similar to \citet{szymanski2023aren}. At a high level \citet{szymanski2023aren} analyzed the relationship between ASR performance and NER model performance. They found that the NER models make errors on the ASR-transcribed data, even when the entity is contained in the transcript. This can be caused by covariate shift (e.g., we would not expect a model trained on general data to generalize to biomedical articles). However, our work differs in three major ways. First, the focus of this paper is on the biomedical domain. There are limited publicly available datasets that researchers can use to develop new methods for improving downstream tasks (e.g., NER) on noisy transcriptions. Second, our focus is on noisy audio. Specifically, our audio contains multiple background speakers and background noise (e.g., TV sounds). Compared to prior work applying NLP to transcripts, our transcripts are not ``clean.'' Third, we introduce a simple method of improving NER system performance without training on domain-specific transcribed data, which is advocated by \citet{szymanski2023aren}. Obtaining NER annotations on noisy transcriptions is time-consuming and infeasible in a timely manner. Hence, our approach can improve existing NER model performance when applied to noisy transcripts with only a few examples of noisy and clean transcripts (the actual NER annotations are not required).

\begin{table*}[t]
\centering
\renewcommand{\arraystretch}{1.35}

\resizebox{\textwidth}{!}{%
\begin{tabular}{lp{15cm}}
\toprule
\textbf{Dataset} & \textbf{Example} \\ \midrule
\multirow{4}{*}{\textbf{CADEC}} & i actually am taking provacal, but when I bring up the drug, it brings me   to lipitor. I have experienced fatigue, hip pain,some joint pain in knee. \\ \cmidrule(lr){2-2}
& I would not recommend this drug,my Doctor didn't explain any risk to taking   this drug,although it lowered my cholesterol some,I changed my diet and   started an exercise plan,I quit taking the drug 2 months ago and have   continually lowered my chol. level. \\ \midrule
\multirow{5.5}{*}{\textbf{Synthetic BTACT}}& Let me see what I can do. mortar, cod, lemming, vole, quail, pigeon, rodent,   laboratory rat strains, turkey breeds, eel, great blue heron, ringneck dove,   bonobo, prawn, record. That's something I'll need some more time to consider.   rodent. I'm concerned that I might not be able to provide a well-informed   response. rodent, laboratory rat strains, turkey breeds, eel, great blue   heron, ringneck dove, bonobo, prawn, pigeon, record, cockroach, pike \\ \cmidrule(lr){2-2}
 & Okay, let's get to work. loquat, mouse melon, soda, kiwifruit, cucumber,   lime, plantain, white currant, mouse melon, height, rambutan, apple,   cucumber, citrus, lime, jackfruit, goji berry, loquat. \\ \bottomrule
\end{tabular}}
\caption{Modified examples from the CADEC and Synthetic BTACT datasets.}
\label{tab:data}
\end{table*}
\section{Data}

This paper uses CADEC~\cite{karimi2015cadec} and a Synthetic BTACT dataset. CADEC is a popular NER dataset for extracting adverse drug reactions from experiences written by patients, and the Synthetic BTACT dataset is a novel dataset we created that simulates questions of the Brief Test of Adult Cognition by Telephone (BTACT). For both datasets, we have research assistants read each item and record an audio file of the reading. We generate noisy audio files by merging the files of multiple speakers and background noises/sounds. A high-level overview of the data collection process is shown in Figure~\ref{fig:dproc}. The details of the curation and creation are described in the following subsections.

\subsection{Dataset Curation}

\paragraph{CADEC.}  The CSIRO Adverse Drug Event Corpus (CADEC)\footnote{The dataset is publicly accessible at https://data.csiro.au.} is an extensively annotated collection of medical forum posts centered on patient-reported Adverse Drug Events (ADEs). Derived from social media discussions, the corpus comprises text predominantly written in colloquial language, often straying from conventional English grammar and punctuation norms. The annotations reference various concepts, including drugs, adverse effects, symptoms, and associated diseases, all linked to controlled vocabularies such as SNOMED Clinical Terms and MedDRA. Rigorous annotation guidelines, multi-stage annotations, inter-annotator agreement assessments, and a final review by a clinical terminologist ensure the high quality of annotations. This corpus, initially sourced from \url{Askapatient.com}, proves invaluable for research in information extraction and broader text mining from social media, especially for identifying potential adverse drug reactions directly reported by patients. This resource empowers patients by encouraging the sharing of side effects and success stories, advocating for informed health decisions through real-life experiences with drug treatments. \url{AskaPatient.com} provides tools to support and inform the engaged patient. Overall, the entity types in the dataset are adverse drug reaction (ADR), drug, finding, disease, and symptom. Examples can be found in Table~\ref{tab:data}.

\begin{table}[t]
\centering
\resizebox{\linewidth}{!}{%
\begin{tabular}{@{}lrrr@{}}
\toprule
 & Text Files &  \# Audio  & \# Types \\ \midrule
CADEC & 1,250 & 1,000 &  5 \\
Syntehtic BTACT & 1,000 & 1,000 & 3  \\ \bottomrule
\end{tabular}%
}
\caption{This table reports the basic dataset statistics for both CADEC and Synthetic BTACT, including the number of audio files (\# Audio) and the number of named entity classes (\# Types).}
\label{tab:my-table}
\end{table}

\begin{figure*}[t]
    \centering
    \includegraphics[width=\textwidth]{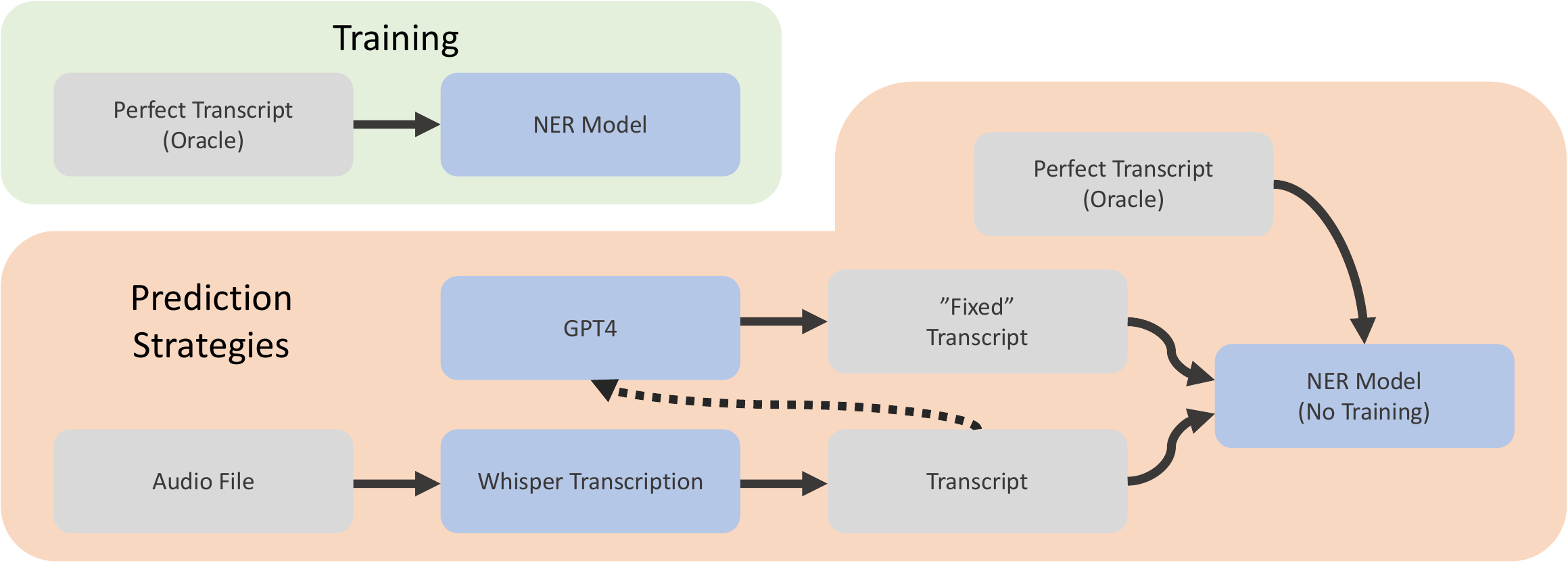}
    \caption{Overview of the training procedure and the prediction strategies we explore to improve biomedical NER performance.}
    \label{fig:enter-label}

\end{figure*}

\paragraph{Synthetic BTACT.} The Brief Test of Adult Cognition by Telephone (BTACT) is a succinct yet comprehensive screening tool designed for assessing cognitive function, particularly in the context of dementia. Administered either in person or over the phone, this battery of tests evaluates key cognitive domains, including episodic verbal memory, working memory, verbal fluency, inductive reasoning, and processing speed. Developed for use in the National Survey of Midlife Development in the United States (MIDUS), the BTACT combines adapted neuropsychological tests with novel subtests. Extensive research has validated its effectiveness as a dementia screening measure across a diverse range of individuals, providing a valuable tool for early detection and intervention in cognitive decline. This versatile assessment tool holds promise for enhancing dementia diagnosis and care, particularly in situations where in-person evaluation may not be feasible.

We create synthetic BTACT subtest answers for the questions, ``List as many animals as possible in 30 seconds'' and ``List as many fruits as possible in 30 seconds.''  Specifically, we randomly generate a list of fruits or animals using publicly available lexicons.\footnote{\url{https://github.com/imsky/wordlists}} Next, we randomly inject incorrect entities (non-animal and non-fruit) into the respective lists. Next, we randomly add an introduction sentence (e.g., ``Okay, let me try to list as many as I can.'') and interjections (e.g., ``Let me think for a second'') in the middle of the lists. The entity types are ``animal'', ``fruit'', and ``other.'' Examples can be found in Table~\ref{tab:data}.

\subsection{Audio Recordings and mixing}

Next, we had research assistants read each of the scripts from both datasets. Luckily, the examples in both datasets are written in first-person, which also helps more natural readings. In total, we had five diverse assistants with respect to age and gender that helped the recording process. Next, each recording was normalized to ensure the volume (loudness) was the same across all speakers using pyloudnorm~\cite{steinmetz2021pyloudnorm}.

After generating audio recordings from both datasets, we randomly sampled the signal-to-interference (SNR) ratio to merge each audio file with the audio files of 2 to 3 other speakers and a background noise/sound. The SNR used for the CADEC and the Synthetic BTACT dataset differs when we generate noisy files. We differ in the SNR ranges because the CADEC dataset has more ``signal'' because of the relative fluency of the text. Intuitively, the transcription models can understand how to extract words when they follow common syntactic patterns (e.g., a noun follows a determiner). However, the Synthetic BTACT data contains large lists of nouns, and the relation of one noun to the next provides little information for prediction. An example of this phenomenon can be seen in Figure~\ref{tab:data}. For the CADEC dataset, we randomly mix each audio file with other users using an SNR sampled from $\{-1,0, 6\}$ (negative scores mean the background is ``louder'' than the main speaker), and the background noise SNR is sampled from $\{-1,0,3,6,9,12\}$. For the Synthetic BTACT dataset, we randomly sample the SNR from the set $\{4, 6, 9\}$, and the background SNR is sampled from $\{3,6,9,12\}$.  The background noise types include kitchen, TV, home appliances, music, and other ambient noises sampled from recordings of "daily life" environments.  Overall, this mixing strategy is based on the work by \citet{ji2020speaker}.


\section{Methods}

This study attempts to investigate the performance of NER systems on biomedical ASR-transcribed data. In this regard, after developing a set of baselines trained on original scripts, we transcribe the noisy audio and evaluate the NER performances on the noisy and original transcripts. 

In addition, we introduce a simple framework to improve the NER system performance that requires no training on domain-specific transcribed data. We use the fourth iteration of the Generative Pre-trained Transformer (GPT4)~\cite{OpenAI2023GPT4TR} to post-process the ASR transcripts using its advanced capacity in contextual understanding and extensive knowledge, which covers a very broad range of biomedical concepts. During this post-processing, GPT4 is provided instructions to evaluate and refine the transcribed outputs to improve the downstream NER performance. In this regard, we study two approaches: zero-shot prompting and few-shot in-context learning.

\begin{figure}[t]
    \centering
    \includegraphics[width=.7\linewidth]{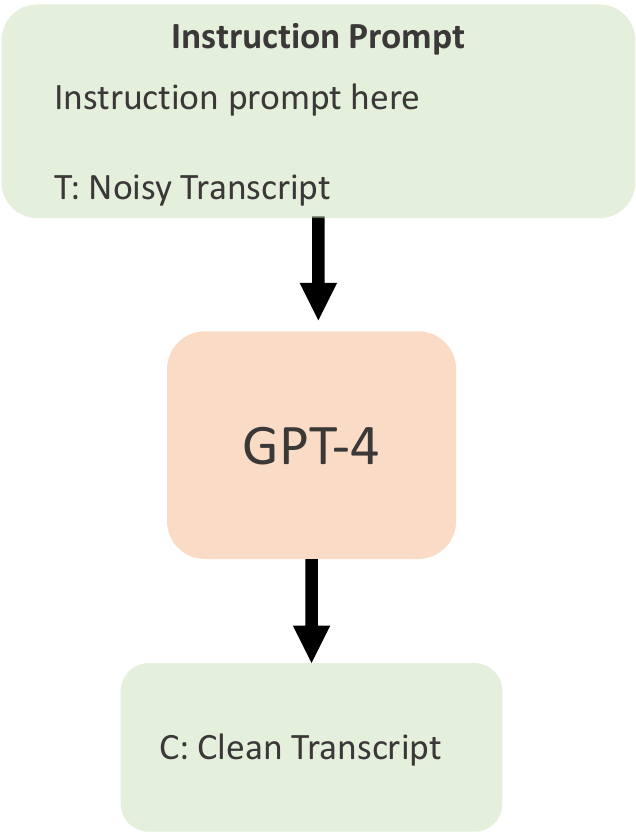}
    \caption{Example of the zero-shot prompting strategy we use for GPT4.}
    \label{fig:zeroshot}

\end{figure}

\paragraph{NER baselines.}
Using the dataset, we develop a set of solid and representative baselines for biomedical NER and evaluate them on original scripts as well as the ASR-transcribed data. However, we use the original scripts for training the models using the widely adopted BIO tagging scheme~\cite{sang2003introduction}. We use pretrained language models such as BERT~\cite{devlin2018bert}, BioBERT~\cite{lee2020biobert}, T5~\cite{raffel2020exploring} and Flair's\footnote{Flair is a framework for many NLP tasks including NER, POS tagging and text classification which provides a variety of embeddings as well as modules to combine with pretrained language models. \url{https://flairnlp.github.io/}} pretrained word embeddings such as GloVe that are fine-tuned on news articles and PubMed datasets. We combine these with the widely adopted BiLSTM-CRF~\cite{lample2016neural, sui2021large}.

\paragraph{ASR.}
As for the ASR module, we use Whisper from OpenAI~\cite{radford2023robust} because of its high-quality transcriptions and wide adoption in the literature~\cite{zhuo2023lyricwhiz}. We use Whisper in an audio streaming fashion in which the audio input is divided into overlapping chunks, and the overlapping content is post-processed using Llama-2-7b~\cite{touvron2023llama} by giving instructions to concatenate the transcribed chunks--we also have tested with GPT4, but there has been a minimal improvement. The size of the chunk is also chosen based on optimum WER.

\paragraph{Zero-shot Prompting.}
Taking advantage of GPT4's understanding of context and comprehensive knowledge in various biomedical domains, we instruct it to refine the transcript, knowing that the end goal is to improve the performance of the downstream NER system evaluated based on detecting the correct terms and categories.

\begin{figure}[t]
    \centering
    \includegraphics[width=\linewidth]{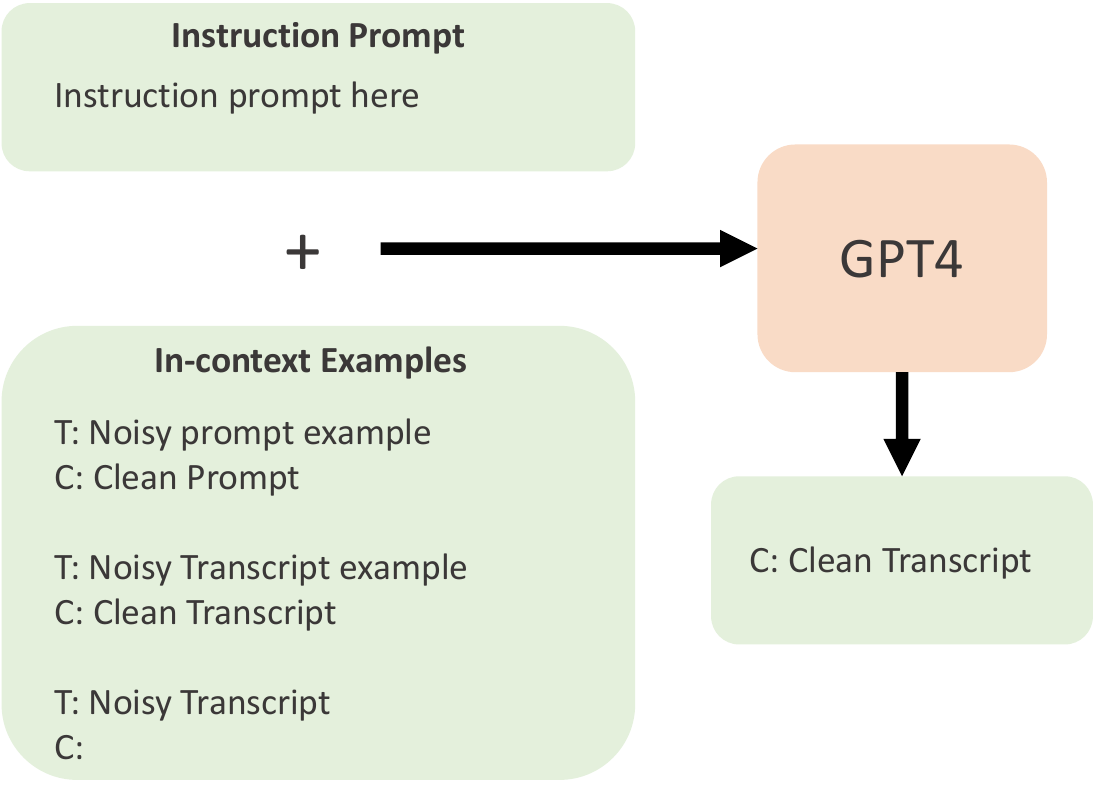}
    \caption{Example of the few-shot prompting strategy we use for GPT4.}
    \label{fig:fewshot}

\end{figure}

The format of the zero-shot prompting is shown in Figure~\ref{fig:zeroshot}. In addition, we provide the following instructions and contextual information: \textbf{i)} We explain the general topic discussed in each dataset; for example, for CADEC, we mention that the data is a transcribed medical conversation about adverse drug reactions. For BTACT dataset, we provide examples of valid animals/fruits. \textbf{ii)} We mention that the audio is noisy, and some words may have been incorrectly transcribed. It has to detect the inappropriate terms and also rephrase them to phonetically similar, yet more appropriate ones. \textbf{iii)} We also explain the multi-speaker nature of the noise and mention the possibility that some words may be transcribed from the background speakers. This way, GPT4 provides a more concise transcript or at least removes the off-topic sentences to improve the performance of the downstream NER.

\paragraph{Few-shot In-Context Learning.}
We also leverage in-context learning to provide sample ASR noisy transcripts alongside the corresponding ground-truth script and tagged named entities. This way, we teach the GPT4 model through direct examples, and it learns to identify the relationships between the errors and correct similar errors in the test transcripts.  We show the prompt format we generally use in Figure~\ref{fig:fewshot}. The GPT4 may also uncover unique relationships and come up with innovative approaches to refine the transcript. By exposing the model to various examples, from common transcription inaccuracies to more complex ones, GPT4 is encouraged to mimic corrections and understand the underlying principles of these mistakes~\cite{ge2022few,gutierrez2022thinking,jin2023genegpt,cheng2023exploring}.


To choose the set of examples, we randomly sample examples from the training set and cluster them based on their type of errors. In this regard, we look at the NER precision, recall, and F1; differences between the transcript and the original script along with unrecognized or misrecognized named entities. We also ask the GPT4 model to provide its insights about what might have caused the errors and how it can fix the errors. This insight is also used in the clustering algorithm. Finally, a set of varied examples is chosen to be provided to our in-context few-shot learning approach\footnote{We have tested the effect of every cluster as well as various combinations, but providing examples from all clusters results in the maximum improvement.}.

\begin{table}[t]
\centering
\resizebox{\linewidth}{!}{
\begin{tabular}{llrrr}
\toprule
 & Model & Precision & Recall & F1 \\ \midrule
\multirow{5}{*}{Original} & BERT & .669 & .569 & .615 \\
 & BioBERT & .665 & .583 & .622\\
 & T5 & .679 & .585 & .629 \\
 & Flair & .663 & .658 & .660 \\ \cmidrule(lr){2-5}
 & Average &  .669 & .599 & .631 \\ \midrule \midrule
\multirow{5}{*}{Whisper}  & BERT & .215 & .215 & .229 \\
 & BioBERT & .243 & .208 & .224 \\
 & T5 & .242 & .272 & .256 \\
 & Flair & .236 & .242 & .239 \\ \cmidrule(lr){2-5}
 & Average &  .234 & .234 & .237\\ \midrule 
\multirow{5}{*}{+GPT4} & BERT & .363 & .336 & .349 \\
& BioBERT & .363 & .347 & .355 \\
& T5 & .362 & .334 & .347 \\
& Flair & .352 & .389 & .369 \\ \cmidrule(lr){2-5}
 & Average & .360 & .351 & .355 \\ \midrule 
\multirow{5}{*}{+GPT4+Few-shot} & BERT & .371 & .358 & .364 \\
 & BioBERT & .387 & .383 & .385\\
& T5 & \textbf{.374} & .360 & .367  \\
& Flair & .367 & \textbf{.415} & \textbf{.389} \\  \cmidrule(lr){2-5}
 & Average &\textbf{.375} & \textbf{.379} & \textbf{.376}\\ \bottomrule
\end{tabular}
}
\caption{CADEC ASR dataset results}
\label{table:cadec}

\end{table}
\section{Results}
In this section, we evaluate and contrast ASR-NER performances on biomedical noisy transcripts with those of the two proposed methods. We evaluate the models based on their performance in detecting the named entities as well as their categories. We use micro precision, recall, and f1 as the evaluation metrics in which a correct prediction happens only if the named entity and the corresponding tag category $\hat{y^{\hat{c_i}}_i}=(\hat{n_i},\hat{c_i})$ is predicted correctly and matches those of the corresponding original ground truth $y^{c_i}_i=(n_i,c_i)$. If a script has multiple pairs of the same named entity and category, we treat each as a separate prediction to account for repeated terms. This way, our evaluation would be very similar to CoNLL's~\cite{sang2003introduction}; however, we do not perform a Span detection or consider the BIO tags as the position of words changes in all of our noisy transcript datasets due to the existence of mistranscribed words. 

Table \ref{table:cadec} shows the performances on the CADEC dataset. As you can see, and was expected, the performance on the NER has significantly dropped (on average 62\% of micro f1 scores) on the ASR noisy transcript data for different reasons, including mistranscribed words/terms, the words that are picked up from the background and those that are missed to name but a few reasons. Another reason is the covariate shift between the training and testing set. For example, the ASR output may write the full name of a drug/disease, but the original script may use abbreviations, or they may use different punctuation. PLM-based models' performances have dropped less due to their robustness to covariate shift and their ability to handle Out-Of-Vocabulary (OOV) scenarios (in comparison with traditional pretrained embeddings such as GloVe). T5 performs consistently better on the recall and precision, which shows its superior robustness, and it would potentially be an ideal choice for fine-tuning if one decides to fine-tune the NER on the noisy transcript data.

\begin{table}[t]
\centering
\resizebox{\linewidth}{!}{
\begin{tabular}{llrrr}
\toprule
Corpus & Model & Precision & Recall & F1 \\ \midrule
\multirow{4}{*}{Original} & BERT & .974 & .972 & .973 \\
 & T5 & .962 & .968 & .965 \\
 & Flair & .942 & .963 & .953 \\ \cmidrule(lr){2-5}
 & AVG & .959 & .968 & .964 \\ \midrule \midrule
\multirow{4}{*}{Whisper} & BERT & .528 & .525 & .526 \\
 & T5 & .554 & .609 & .580 \\
 & Flair & .555 & .624 & .587 \\ \cmidrule(lr){2-5}
 & AVG & .546 & .586 & .564 \\ \midrule
\multirow{4}{*}{+GPT4} & BERT & .554 & .584 & .570 \\
 & T5 & .571 & .609 & .589 \\
 & Flair & .580 & .614 & .597 \\ \cmidrule(lr){2-5}
 & AVG & .568 & .602 & .585 \\ \midrule
\multirow{4}{*}{+GPT4+Few-shot} & BERT & .567 & .598 & .584 \\
 & T5 & .611 & \textbf{.659} & .634 \\
 & Flair & \textbf{.620} & .649 & \textbf{.634} \\ \cmidrule(lr){2-5}
 & AVG & .599 & .635 & \textbf{.617} \\ \bottomrule
\end{tabular}%
}
\caption{Overall results on the Synthetic BTACT ASR dataset. }
\label{table:btact}
\end{table}

The \textbf{zero-shot prompting} consistently improves the micro f1 score by an average of 0.14 across all models, which is 59\% improvement, fixing the 22\% of the drop (reduces the 62\% drop to 40\%). This shows the value and effectiveness of providing the context and GPT4's capability of utilizing its knowledge to refine the transcripts. With the improvement of the transcripts and reduction in noise models, performances improved, and Flair pretrained embedding surpassed other models, consistent with the original scripts' results. BioBERT and Flair improved more than others, especially T5, because GPT4 may replace some terms with new medical terms that T5 has not seen in the training set, but BioBERT and Flair PubMed embeddings are good at recognizing them. 

The \textbf{few-shot in-context learning} additionally improves the micro f1 score by an average of 0.03 across all models. This is 79\% improvement from the ASR noisy transcripts that happen consistently across all models and fixes the f1 drop by 27\%. Furthermore, the difference between the performance of the models is more similar to those of the original scripts, even in comparison with the zero-shot prompting, which is because the post-transcribed scripts are more similar to the original ones. However, the BioBERT model, which is the best-performing model, has shown the most improvement by improving by 32\% in terms of micro-F1. Flair and BioBERT are the best models for zero-shot and few-shot learning when used with GPT4 because of their pretraining on medical terms. We hypothesize that more specialized language models that use medical databases for pretraining can potentially improve performance.

Table \ref{table:btact} shows the performances on the Synthetic BTACT dataset. Similar to the CADEC dataset, we see that the performance significantly drops on the noisy transcripts, and zero- or few-shot learning will consistently improve the performance. The generalization of improvements across the two datasets suggests the effectiveness of the proposed approaches. However, the improvement on BTACT data is substantially lower than CADEC--on BTACT data, the micro-f1 drop reduces from 41.4\% in the noisy ASR-NER to 35.9\% using few-shot learning, i.e., 5.5\% improvement. This is due to the unnatural style of the BTACT data and the lack of context to be leveraged by GPT4.

\section{Discussion and Error Analysis}
The error in ASR-NER can come from different sources, such as noise in the original script, NER errors, or ASR errors in which words may be mistranscribed or not transcribed from the background speaker by mistake. The noise from the original script includes grammatical errors, typos, or even biomedical misconceptions. Although in a general setting, we cannot address the input data noise, in an ASR setting, this causes the model to either mistranscribe or fix the issue, which causes a mismatch between the predictions and ground-truth tags. For example, ('flu like symtoms', 'ADR') is transcribed to ('flu-like symtoms', 'ADR'). Another example, if the script talks about a drug that can stop multiple episodes of migraines, it should say
\begin{quote}
    ``\textit{It stops migraine\textbf{s}}''
\end{quote}
rather than
\begin{quote}
    ``\textit{It stops migraine}.'' 
\end{quote}
GPT4 realizes these issues and fixes them using its biomedical knowledge.

The errors that come from the NER system cannot be remedied, but sometimes the noisy transcribed data can add terminologies or a combination of words, which results in errors in the NER model; for example, NER is prone to misidentify pronouns as ADRs, e.g. (we, ADR) or (i, ADR), especially when there are not many named-entities in the input text. Therefore, when the input is very noisy, and ASR cannot transcribe any ADR terms, the output includes many such errors. Furthermore, the covariate shift between the ASR transcript and the original script can also cause errors in the NER system. However, both types of errors are mitigated to some extent by few-shot in-context learning as the GPT4 has seen these errors and changes in the style of the transcript compared to the original script. 

The error can also come from Whisper's limited vocabulary or noisy predictions during uncertainty. For example, Whisper cannot detect many drug names, especially phonetically similar to a more common term. For example, it transcribes Arthotec as ``arthritis'' or ``arthrotype,'' but providing the context that the script is about adverse drug effects fixes the problem. However, due to the intrinsic randomness of GPT4, there are examples in which the term is not fixed in either zero- or few-shot learning--this happens for many other terms, and very commonly, one approach has the right answer, which suggests the use of ensemble approaches. 

We also find unique situations where GPT4 hallucinates new contexts. For instance, The original text 
\begin{quote}
    ``\textit{Both husband and wife on a low dosage (10 mg). We are having extreme reactions to heat.}''
\end{quote} is transcribed by Whisper to
\begin{quote}
    ``\textit{There you have it everyone, some kind of low dose of temperature, we are having extreme reactions}.''.
\end{quote}
But, after applying GPT4, it hallucinates a change in temperature, e.g.,
\begin{quote}
    ``\textit{We both have extreme reactions to temperature changes}.''
\end{quote}The new context has unexpected impacts on the NER models, e.g., ``changes in temperature'' is detected as an ADR. Future research is required to reduce these hallucinations.



\section{Conclusion}

This paper explores the ASR-NLP gap in the biomedical domain, particularly for noisy audio. This challenge is especially pronounced in biomedical Named Entity Recognition (NER) tasks. While advancements in ASR show promise in controlled environments, real-world noisy conditions present significant obstacles (e.g., lack of publicly available datasets). To address this, we've introduced the BioASR-NER dataset, offering a mix of clean and noisy biomedical recordings. Coupled with our innovative GPT4-based transcript cleaning approach, we've made strides toward mitigating transcription errors.

There are two avenues for future research. First, our methodology to fix transcripts is based on the transcription text. Incorporating the audio information, particularly with recent advances in transformer-based audio-representations~\cite{gong2022ssast}, could substantially improve performance. Second, other biomedical NLP-based tasks, particularly when applied to noisy transcripts, deserve attention. These tasks include tasks such as text summarization~\cite{mishra2014text} and question answering~\cite{jin2022biomedical}. How these models generalize when applied to audio-generated transcripts with background voices and noises is unclear. Hence, we will explore these tasks as future work.

\section*{Acknowledgements}
This material is based upon work supported by
the National Science Foundation (NSF) under
Grant No. 2145357.

\nocite{*}
\section{Bibliographical References}\label{sec:reference}

\bibliographystyle{lrec-coling2024-natbib}
\bibliography{lrec-coling2024-example}

\begin{thebibliography}{50}
\expandafter\ifx\csname natexlab\endcsname\relax\def\natexlab#1{#1}\fi

\bibitem[{Chen et~al.(2023)Chen, Wang, Lin, Zhao, and Yang}]{chen2023few}
Peng Chen, Jian Wang, Hongfei Lin, Di~Zhao, and Zhihao Yang. 2023.
\newblock Few-shot biomedical named entity recognition via knowledge-guided
  instance generation and prompt contrastive learning.
\newblock \emph{Bioinformatics}, 39(8):btad496.

\bibitem[{Cheng et~al.(2023)Cheng, Guo, He, Lu, Gu, and
  Wu}]{cheng2023exploring}
Kunming Cheng, Qiang Guo, Yongbin He, Yanqiu Lu, Shuqin Gu, and Haiyang Wu.
  2023.
\newblock Exploring the potential of gpt-4 in biomedical engineering: the dawn
  of a new era.
\newblock \emph{Annals of Biomedical Engineering}, pages 1--9.

\bibitem[{Chiu et~al.(2021)Chiu, Yeh, Lin, and Chang}]{chiu2021recognizing}
Yu-Wen Chiu, Wen-Chao Yeh, Sheng-Jie Lin, and Yung-Chun Chang. 2021.
\newblock Recognizing chemical entity in biomedical literature using a
  bert-based ensemble learning methods for the biocreative 2021 nlm-chem track.
\newblock In \emph{Proceedings of the seventh BioCreative challenge evaluation
  workshop}.

\bibitem[{Devlin et~al.(2018)Devlin, Chang, Lee, and
  Toutanova}]{devlin2018bert}
Jacob Devlin, Ming-Wei Chang, Kenton Lee, and Kristina Toutanova. 2018.
\newblock Bert: Pre-training of deep bidirectional transformers for language
  understanding.
\newblock \emph{arXiv preprint arXiv:1810.04805}.

\bibitem[{Dhuriya et~al.(2022)Dhuriya, Chadha, Gupta, Shah, Chhimwal, Gaur, and
  Raghavan}]{dhuriya2022improving}
Ankur Dhuriya, Harveen~Singh Chadha, Anirudh Gupta, Priyanshi Shah, Neeraj
  Chhimwal, Rishabh Gaur, and Vivek Raghavan. 2022.
\newblock Improving speech recognition for indic languages using language
  model.
\newblock \emph{arXiv preprint arXiv:2203.16595}.

\bibitem[{Ganoe et~al.(2021)Ganoe, Wu, Barr, Haslett, Dannenberg, Bonasia,
  Finora, Schoonmaker, Onsando, Ryan et~al.}]{ganoe2021natural}
Craig~H Ganoe, Weiyi Wu, Paul~J Barr, William Haslett, Michelle~D Dannenberg,
  Kyra~L Bonasia, James~C Finora, Jesse~A Schoonmaker, Wambui~M Onsando, James
  Ryan, et~al. 2021.
\newblock Natural language processing for automated annotation of medication
  mentions in primary care visit conversations.
\newblock \emph{JAMIA open}, 4(3):ooab071.

\bibitem[{Ge et~al.(2022)Ge, Guo, Yang, Al-Garadi, and Sarker}]{ge2022few}
Yao Ge, Yuting Guo, Yuan-Chi Yang, Mohammed~Ali Al-Garadi, and Abeed Sarker.
  2022.
\newblock Few-shot learning for medical text: A systematic review.
\newblock \emph{arXiv preprint arXiv:2204.14081}.

\bibitem[{Ghosh et~al.(2023)Ghosh, Tyagi, Kumar, and Manocha}]{ghosh2023bioaug}
Sreyan Ghosh, Utkarsh Tyagi, Sonal Kumar, and Dinesh Manocha. 2023.
\newblock Bioaug: Conditional generation based data augmentation for
  low-resource biomedical ner.
\newblock \emph{arXiv preprint arXiv:2305.10647}.

\bibitem[{Gong et~al.(2022)Gong, Lai, Chung, and Glass}]{gong2022ssast}
Yuan Gong, Cheng-I Lai, Yu-An Chung, and James Glass. 2022.
\newblock Ssast: Self-supervised audio spectrogram transformer.
\newblock In \emph{Proceedings of the AAAI Conference on Artificial
  Intelligence}, volume~36, pages 10699--10709.

\bibitem[{Guan and Zhou(2023)}]{guan2023prefix}
Zhengyi Guan and Xiaobing Zhou. 2023.
\newblock A prefix and attention map discrimination fusion guided attention for
  biomedical named entity recognition.
\newblock \emph{BMC bioinformatics}, 24(1):42.

\bibitem[{Gutierrez et~al.(2022)Gutierrez, McNeal, Washington, Chen, Li, Sun,
  and Su}]{gutierrez2022thinking}
Bernal~Jimenez Gutierrez, Nikolas McNeal, Clay Washington, You Chen, Lang Li,
  Huan Sun, and Yu~Su. 2022.
\newblock Thinking about gpt-3 in-context learning for biomedical ie? think
  again.
\newblock \emph{arXiv preprint arXiv:2203.08410}.

\bibitem[{Hacking et~al.(2023)Hacking, Verbeek, Hamers, and
  Aarts}]{hacking2023development}
Coen Hacking, Hilde Verbeek, Jan~PH Hamers, and Sil Aarts. 2023.
\newblock The development of an automatic speech recognition model using
  interview data from long-term care for older adults.
\newblock \emph{Journal of the American Medical Informatics Association},
  30(3):411--417.

\bibitem[{Ji et~al.(2020)Ji, Yu, Zhang, Su, Yu, Liu, and Yu}]{ji2020speaker}
Xuan Ji, Meng Yu, Chunlei Zhang, Dan Su, Tao Yu, Xiaoyu Liu, and Dong Yu. 2020.
\newblock Speaker-aware target speaker enhancement by jointly learning with
  speaker embedding extraction.
\newblock In \emph{ICASSP 2020-2020 IEEE International Conference on Acoustics,
  Speech and Signal Processing (ICASSP)}, pages 7294--7298. IEEE.

\bibitem[{Jia et~al.(2019)Jia, Liang, and Zhang}]{jia2019cross}
Chen Jia, Xiaobo Liang, and Yue Zhang. 2019.
\newblock Cross-domain ner using cross-domain language modeling.
\newblock In \emph{Proceedings of the 57th annual meeting of the association
  for computational linguistics}, pages 2464--2474.

\bibitem[{Jin et~al.(2023)Jin, Yang, Chen, and Lu}]{jin2023genegpt}
Qiao Jin, Yifan Yang, Qingyu Chen, and Zhiyong Lu. 2023.
\newblock Genegpt: Augmenting large language models with domain tools for
  improved access to biomedical information.
\newblock \emph{ArXiv}.

\bibitem[{Jin et~al.(2022)Jin, Yuan, Xiong, Yu, Ying, Tan, Chen, Huang, Liu,
  and Yu}]{jin2022biomedical}
Qiao Jin, Zheng Yuan, Guangzhi Xiong, Qianlan Yu, Huaiyuan Ying, Chuanqi Tan,
  Mosha Chen, Songfang Huang, Xiaozhong Liu, and Sheng Yu. 2022.
\newblock Biomedical question answering: a survey of approaches and challenges.
\newblock \emph{ACM Computing Surveys (CSUR)}, 55(2):1--36.

\bibitem[{Karimi et~al.(2015)Karimi, Metke-Jimenez, Kemp, and
  Wang}]{karimi2015cadec}
Sarvnaz Karimi, Alejandro Metke-Jimenez, Madonna Kemp, and Chen Wang. 2015.
\newblock Cadec: A corpus of adverse drug event annotations.
\newblock \emph{Journal of biomedical informatics}, 55:73--81.

\bibitem[{King et~al.(2023)King, Angus, Cooper, Mowery, Seaman, Potter,
  Bukowski, Al-Khafaji, Gunn, and Kahn}]{king2023voice}
Andrew~J King, Derek~C Angus, Gregory~F Cooper, Danielle~L Mowery, Jennifer~B
  Seaman, Kelly~M Potter, Leigh~A Bukowski, Ali Al-Khafaji, Scott~R Gunn, and
  Jeremy~M Kahn. 2023.
\newblock A voice-based digital assistant for intelligent prompting of
  evidence-based practices during icu rounds.
\newblock \emph{Journal of Biomedical Informatics}, 146:104483.

\bibitem[{Kodish-Wachs et~al.(2018)Kodish-Wachs, Agassi, Kenny~III, and
  Overhage}]{kodish2018systematic}
Jodi Kodish-Wachs, Emin Agassi, Patrick Kenny~III, and J~Marc Overhage. 2018.
\newblock A systematic comparison of contemporary automatic speech recognition
  engines for conversational clinical speech.
\newblock In \emph{AMIA Annual Symposium Proceedings}, volume 2018, page 683.
  American Medical Informatics Association.

\bibitem[{Lample et~al.(2016)Lample, Ballesteros, Subramanian, Kawakami, and
  Dyer}]{lample2016neural}
Guillaume Lample, Miguel Ballesteros, Sandeep Subramanian, Kazuya Kawakami, and
  Chris Dyer. 2016.
\newblock Neural architectures for named entity recognition.
\newblock \emph{arXiv preprint arXiv:1603.01360}.

\bibitem[{Leaman and Gonzalez(2008)}]{leaman2008banner}
Robert Leaman and Graciela Gonzalez. 2008.
\newblock Banner: an executable survey of advances in biomedical named entity
  recognition.
\newblock In \emph{Biocomputing 2008}, pages 652--663. World Scientific.

\bibitem[{Lee et~al.(2020)Lee, Yoon, Kim, Kim, Kim, So, and
  Kang}]{lee2020biobert}
Jinhyuk Lee, Wonjin Yoon, Sungdong Kim, Donghyeon Kim, Sunkyu Kim, Chan~Ho So,
  and Jaewoo Kang. 2020.
\newblock Biobert: a pre-trained biomedical language representation model for
  biomedical text mining.
\newblock \emph{Bioinformatics}, 36(4):1234--1240.

\bibitem[{L{\'o}pez-{\'U}beda et~al.(2021)L{\'o}pez-{\'U}beda,
  D{\'\i}az-Galiano, Ure{\~n}a-L{\'o}pez, and
  Mart{\'\i}n-Valdivia}]{lopez2021combining}
Pilar L{\'o}pez-{\'U}beda, Manuel~Carlos D{\'\i}az-Galiano, L~Alfonso
  Ure{\~n}a-L{\'o}pez, and M~Teresa Mart{\'\i}n-Valdivia. 2021.
\newblock Combining word embeddings to extract chemical and drug entities in
  biomedical literature.
\newblock \emph{BMC bioinformatics}, 22(1):1--18.

\bibitem[{Mani et~al.(2020)Mani, Palaskar, and Konam}]{mani2020towards}
Anirudh Mani, Shruti Palaskar, and Sandeep Konam. 2020.
\newblock Towards understanding asr error correction for medical conversations.
\newblock In \emph{Proceedings of the first workshop on natural language
  processing for medical conversations}, pages 7--11.

\bibitem[{Mishra et~al.(2014)Mishra, Bian, Fiszman, Weir, Jonnalagadda,
  Mostafa, and Del~Fiol}]{mishra2014text}
Rashmi Mishra, Jiantao Bian, Marcelo Fiszman, Charlene~R Weir, Siddhartha
  Jonnalagadda, Javed Mostafa, and Guilherme Del~Fiol. 2014.
\newblock Text summarization in the biomedical domain: a systematic review of
  recent research.
\newblock \emph{Journal of biomedical informatics}, 52:457--467.

\bibitem[{Nguyen et~al.(2022)Nguyen, Du, Buntine, Chen, and
  Beare}]{nguyen2022hardness}
Ngoc~Dang Nguyen, Lan Du, Wray Buntine, Changyou Chen, and Richard Beare. 2022.
\newblock Hardness-guided domain adaptation to recognise biomedical named
  entities under low-resource scenarios.
\newblock In \emph{Proceedings of the 2022 Conference on Empirical Methods in
  Natural Language Processing}, pages 4063--4071.

\bibitem[{OpenAI(2023)}]{OpenAI2023GPT4TR}
OpenAI. 2023.
\newblock Gpt-4 technical report.
\newblock \emph{ArXiv}, abs/2303.08774.

\bibitem[{Paats et~al.(2015)Paats, Alum{\"a}e, Meister, and
  Fridolin}]{paats2015evaluation}
A~Paats, T~Alum{\"a}e, E~Meister, and I~Fridolin. 2015.
\newblock Evaluation of automatic speech recognition prototype for estonian
  language in radiology domain: a pilot study.
\newblock In \emph{16th Nordic-Baltic Conference on Biomedical Engineering: 16.
  NBC \& 10. MTD 2014 joint conferences. October 14-16, 2014, Gothenburg,
  Sweden}, pages 96--99. Springer.

\bibitem[{Poerner et~al.(2020)Poerner, Waltinger, and
  Sch{\"u}tze}]{poerner2020inexpensive}
Nina Poerner, Ulli Waltinger, and Hinrich Sch{\"u}tze. 2020.
\newblock Inexpensive domain adaptation of pretrained language models: Case
  studies on biomedical ner and covid-19 qa.
\newblock In \emph{Findings of the Association for Computational Linguistics:
  EMNLP 2020}, pages 1482--1490.

\bibitem[{Pyysalo et~al.(2007)Pyysalo, Ginter, Heimonen, Bj{\"o}rne, Boberg,
  J{\"a}rvinen, and Salakoski}]{pyysalo2007bioinfer}
Sampo Pyysalo, Filip Ginter, Juho Heimonen, Jari Bj{\"o}rne, Jorma Boberg,
  Jouni J{\"a}rvinen, and Tapio Salakoski. 2007.
\newblock Bioinfer: a corpus for information extraction in the biomedical
  domain.
\newblock \emph{BMC bioinformatics}, 8:1--24.

\bibitem[{Quiroz et~al.(2019)Quiroz, Laranjo, Kocaballi, Berkovsky,
  Rezazadegan, and Coiera}]{quiroz2019challenges}
Juan~C Quiroz, Liliana Laranjo, Ahmet~Baki Kocaballi, Shlomo Berkovsky, Dana
  Rezazadegan, and Enrico Coiera. 2019.
\newblock Challenges of developing a digital scribe to reduce clinical
  documentation burden.
\newblock \emph{NPJ digital medicine}, 2(1):114.

\bibitem[{Radford et~al.(2023)Radford, Kim, Xu, Brockman, McLeavey, and
  Sutskever}]{radford2023robust}
Alec Radford, Jong~Wook Kim, Tao Xu, Greg Brockman, Christine McLeavey, and
  Ilya Sutskever. 2023.
\newblock Robust speech recognition via large-scale weak supervision.
\newblock In \emph{International Conference on Machine Learning}, pages
  28492--28518. PMLR.

\bibitem[{Raffel et~al.(2020)Raffel, Shazeer, Roberts, Lee, Narang, Matena,
  Zhou, Li, and Liu}]{raffel2020exploring}
Colin Raffel, Noam Shazeer, Adam Roberts, Katherine Lee, Sharan Narang, Michael
  Matena, Yanqi Zhou, Wei Li, and Peter~J Liu. 2020.
\newblock Exploring the limits of transfer learning with a unified text-to-text
  transformer.
\newblock \emph{The Journal of Machine Learning Research}, 21(1):5485--5551.

\bibitem[{Rios et~al.(2018)Rios, Kavuluru, and Lu}]{rios2018generalizing}
Anthony Rios, Ramakanth Kavuluru, and Zhiyong Lu. 2018.
\newblock Generalizing biomedical relation classification with neural
  adversarial domain adaptation.
\newblock \emph{Bioinformatics}, 34(17):2973--2981.

\bibitem[{Rockt{\"a}schel et~al.(2012)Rockt{\"a}schel, Weidlich, and
  Leser}]{rocktaschel2012chemspot}
Tim Rockt{\"a}schel, Michael Weidlich, and Ulf Leser. 2012.
\newblock Chemspot: a hybrid system for chemical named entity recognition.
\newblock \emph{Bioinformatics}, 28(12):1633--1640.

\bibitem[{Sang and De~Meulder(2003)}]{sang2003introduction}
Erik~F Sang and Fien De~Meulder. 2003.
\newblock Introduction to the conll-2003 shared task: Language-independent
  named entity recognition.
\newblock \emph{arXiv preprint cs/0306050}.

\bibitem[{Song et~al.(2021)Song, Li, Liu, and Zeng}]{song2021deep}
Bosheng Song, Fen Li, Yuansheng Liu, and Xiangxiang Zeng. 2021.
\newblock Deep learning methods for biomedical named entity recognition: a
  survey and qualitative comparison.
\newblock \emph{Briefings in Bioinformatics}, 22(6):bbab282.

\bibitem[{Steinmetz and Reiss(2021)}]{steinmetz2021pyloudnorm}
Christian~J. Steinmetz and Joshua~D. Reiss. 2021.
\newblock pyloudnorm: {A} simple yet flexible loudness meter in python.
\newblock In \emph{150th AES Convention}.

\bibitem[{Sui et~al.(2021)Sui, Tian, Chen, Liu, and Zhao}]{sui2021large}
Dianbo Sui, Zhengkun Tian, Yubo Chen, Kang Liu, and Jun Zhao. 2021.
\newblock A large-scale chinese multimodal ner dataset with speech clues.
\newblock In \emph{Proceedings of the 59th Annual Meeting of the Association
  for Computational Linguistics and the 11th International Joint Conference on
  Natural Language Processing (Volume 1: Long Papers)}, pages 2807--2818.

\bibitem[{Sun et~al.(2021)Sun, Yang, Wang, Zhang, Lin, and Wang}]{sun2021deep}
Cong Sun, Zhihao Yang, Lei Wang, Yin Zhang, Hongfei Lin, and Jian Wang. 2021.
\newblock Deep learning with language models improves named entity recognition
  for pharmaconer.
\newblock \emph{BMC bioinformatics}, 22(1):1--16.

\bibitem[{Szyma{\'n}ski et~al.(2023)Szyma{\'n}ski, Augustyniak, Morzy,
  Szymczak, Surdyk, and {\.Z}elasko}]{szymanski2023aren}
Piotr Szyma{\'n}ski, Lukasz Augustyniak, Mikolaj Morzy, Adrian Szymczak,
  Krzysztof Surdyk, and Piotr {\.Z}elasko. 2023.
\newblock Why aren’t we ner yet? artifacts of asr errors in named entity
  recognition in spontaneous speech transcripts.
\newblock In \emph{Proceedings of the 61st Annual Meeting of the Association
  for Computational Linguistics (Volume 1: Long Papers)}, pages 1746--1761.

\bibitem[{Tong et~al.(2021)Tong, Chen, and Shi}]{tong2021multi}
Yiqi Tong, Yidong Chen, and Xiaodong Shi. 2021.
\newblock A multi-task approach for improving biomedical named entity
  recognition by incorporating multi-granularity information.
\newblock In \emph{Findings of the Association for Computational Linguistics:
  ACL-IJCNLP 2021}, pages 4804--4813.

\bibitem[{Touvron et~al.(2023)Touvron, Martin, Stone, Albert, Almahairi,
  Babaei, Bashlykov, Batra, Bhargava, Bhosale et~al.}]{touvron2023llama}
Hugo Touvron, Louis Martin, Kevin Stone, Peter Albert, Amjad Almahairi, Yasmine
  Babaei, Nikolay Bashlykov, Soumya Batra, Prajjwal Bhargava, Shruti Bhosale,
  et~al. 2023.
\newblock Llama 2: Open foundation and fine-tuned chat models.
\newblock \emph{arXiv preprint arXiv:2307.09288}.

\bibitem[{Tran et~al.(2023)Tran, Latif, Reynolds, Park, Elston~Lafata,
  Tai-Seale, and Zheng}]{tran2023mm}
Brian~D Tran, Kareem Latif, Tera~L Reynolds, Jihyun Park, Jennifer
  Elston~Lafata, Ming Tai-Seale, and Kai Zheng. 2023.
\newblock “mm-hm,”“uh-uh”: are non-lexical conversational sounds deal
  breakers for the ambient clinical documentation technology?
\newblock \emph{Journal of the American Medical Informatics Association},
  30(4):703--711.

\bibitem[{van Buchem et~al.(2021)van Buchem, Boosman, Bauer, Kant, Cammel, and
  Steyerberg}]{van2021digital}
Marieke~M van Buchem, Hileen Boosman, Martijn~P Bauer, Ilse~MJ Kant, Simone~A
  Cammel, and Ewout~W Steyerberg. 2021.
\newblock The digital scribe in clinical practice: a scoping review and
  research agenda.
\newblock \emph{NPJ digital medicine}, 4(1):57.

\bibitem[{Vu et~al.(2020)Vu, Phung, and Haffari}]{vu2020effective}
Thuy Vu, Dinh Phung, and Gholamreza Haffari. 2020.
\newblock Effective unsupervised domain adaptation with adversarially trained
  language models.
\newblock In \emph{Proceedings of the 2020 Conference on Empirical Methods in
  Natural Language Processing (EMNLP)}, pages 6163--6173.

\bibitem[{Watanabe et~al.(2022)Watanabe, Ichikawa, Tamura, Iwakura, Ma, and
  Kato}]{watanabe2022auxiliary}
Taiki Watanabe, Tomoya Ichikawa, Akihiro Tamura, Tomoya Iwakura, Chunpeng Ma,
  and Tsuneo Kato. 2022.
\newblock Auxiliary learning for named entity recognition with multiple
  auxiliary biomedical training data.
\newblock In \emph{Proceedings of the 21st Workshop on Biomedical Language
  Processing}, pages 130--139.

\bibitem[{Weber et~al.(2021)Weber, S{\"a}nger, M{\"u}nchmeyer, Habibi, Leser,
  and Akbik}]{weber2021hunflair}
Leon Weber, Mario S{\"a}nger, Jannes M{\"u}nchmeyer, Maryam Habibi, Ulf Leser,
  and Alan Akbik. 2021.
\newblock Hunflair: an easy-to-use tool for state-of-the-art biomedical named
  entity recognition.
\newblock \emph{Bioinformatics}, 37(17):2792--2794.

\bibitem[{Whetten and Kennington(2023)}]{whetten2023evaluating}
Ryan Whetten and Casey Kennington. 2023.
\newblock Evaluating and improving automatic speech recognition using severity.
\newblock In \emph{The 22nd Workshop on Biomedical Natural Language Processing
  and BioNLP Shared Tasks}, pages 79--91.

\bibitem[{Zhuo et~al.(2023)Zhuo, Yuan, Pan, Ma, LI, Zhang, Liu, Dannenberg, Fu,
  Lin et~al.}]{zhuo2023lyricwhiz}
Le~Zhuo, Ruibin Yuan, Jiahao Pan, Yinghao Ma, Yizhi LI, Ge~Zhang, Si~Liu, Roger
  Dannenberg, Jie Fu, Chenghua Lin, et~al. 2023.
\newblock Lyricwhiz: Robust multilingual zero-shot lyrics transcription by
  whispering to chatgpt.
\newblock \emph{arXiv preprint arXiv:2306.17103}.

\end{thebibliography}

\section{Language Resource References}
\label{lr:ref}
\bibliographystylelanguageresource{lrec-coling2024-natbib}
\bibliographylanguageresource{languageresource}

\appendix

\section{Complete System Prompts}

In our study, we structure the dialogues with GPT-4 to refine noisy transcriptions of animal names. The dialogue is initiated with a system prompt, followed by a series of user and assistant interactions. Below, we show the system prompt used for the Synthetic BTACT dataset:
\begin{quote}
    \textbf{System Prompt:} 
"I give you a transcript about animals. Animal names are being called out. 
There are some names that are transcribed by mistake. Fix them to the 
phonetically similar, more proper version if you find it not proper. 
Respond with the fixed transcript only! Remember to remove repetitive 
statements to make the content concise."
\end{quote}

For the CADEC dataset, we use the following system prompt:
\begin{quote}
    \textbf{System Prompt:} "I give you a transcript of medical conversation. There are some technical terms that are transcribed by mistake. Fix them to the phonetically similar, more proper version if you find it not proper. Respond with the fixed transcript only!"
\end{quote}

\end{document}